\documentclass{article}

\usepackage[preprint, nonatbib]{nips_2018}

\usepackage{graphicx}
\usepackage[dvipsnames]{xcolor}
\usepackage{color}
\usepackage{array}
\newcolumntype{?}{!{\vrule width 1pt}}
\usepackage[symbol]{footmisc}

\usepackage[utf8]{inputenc} 
\usepackage[T1]{fontenc}    
\usepackage{url}            
\usepackage{booktabs}       
\usepackage{amsfonts}       
\usepackage{nicefrac}       
\usepackage{microtype}      
\usepackage{amsmath}

\usepackage{bbm}
\usepackage{comment}
\usepackage{multicol}
\usepackage{appendix}
\usepackage{float}
\usepackage{threeparttable}

\newcommand\argmax{\mathrm{argmax}}

\newcommand{\cae}{\operatorname{c_{ae}}}

\newcommand{\norm}[1]{\left\lVert#1\right\rVert}
\newcommand{\sg}[1]{\operatorname{sg}\left(#1\right)}

\newcommand\target{\ensuremath{x}}

\newcommand{\encoder}{z_e}

\newcommand{\multinomial}{\operatorname{Multinomial}}

\title{Theory and Experiments on \\ Vector Quantized Autoencoders}
\author{
  Aurko Roy \texttt{aurkor@google.com} \thanks{Equal contribution} \\
  \And 
  Ashish Vaswani \texttt{avaswani@google.com} \footnotemark[1]\\
  \And 
  Arvind Neelakantan \texttt{aneelakantan@google.com} \footnotemark[1]
  \And 
  Niki Parmar \texttt{nikip@google.com} \footnotemark[1]
}
\begin{document}

\maketitle

\begin{abstract}
Deep neural networks with discrete latent variables offer the promise of better symbolic reasoning, and learning abstractions that are more useful to new tasks. There has been a surge in interest in discrete latent variable models, however, despite several 
recent improvements, the training of discrete latent variable models has remained  challenging and their 
performance has mostly failed to match their continuous counterparts. 
Recent work on vector quantized autoencoders (VQ-VAE) has made substantial progress in this direction,
with its perplexity almost matching that of a VAE on datasets such as CIFAR-10. In this work, 
we investigate an alternate training technique for VQ-VAE, inspired by its connection to the Expectation 
Maximization (EM) algorithm. 
Training the discrete bottleneck with EM helps us achieve better image generation results on CIFAR-10, and 
together with knowledge distillation, allows us to develop a non-autoregressive machine translation model
whose accuracy almost matches a strong greedy autoregressive baseline Transformer,
while being \(3.3\) times faster at inference.
\end{abstract}

\section{Introduction}
Unsupervised learning of meaningful representations is a fundamental problem in machine learning since obtaining labeled data can often be very expensive. 
Continuous representations have largely been the workhorse of unsupervised deep learning models of images \cite{goodfellow2014generative, van2016conditional,
kingma2016improved, salimans2017pixelcnn++, imagetrans}, audio \cite{van2016wavenet, reed2017parallel}, 
and video \cite{kalchbrenner2016video}.
However, it is often the case that datasets are more naturally modeled as a sequence of discrete symbols rather than continuous ones. For example, language and speech are inherently discrete in nature and images are often concisely described by language, see e.g., \cite{vinyals2015show}. 
Improved discrete latent variable models could also prove useful for learning novel data compression algorithms \cite{theis2017lossy}, while having far more interpretable representations of the data.

We build on Vector Quantized Variational Autoencoder (VQ-VAE) \cite{vqvae}, a recently proposed training 
technique for learning discrete latent variables. 
The method uses a learned code-book combined with nearest neighbor search to train the discrete latent variable 
model. 
The nearest neighbor search is performed between the encoder output and the embedding of the latent code using the \(\ell_2\) distance metric. 
The generative process begins by sampling a sequence of discrete latent codes from an autoregressive model fitted on the encoder latents, acting as a learned prior. 
The discrete latent sequence is then consumed by the decoder to generate data. 
The resulting discrete autoencoder obtains impressive results on uncoditional image, speech, and video 
generation. 
In particular, on image generation the performance is almost
on par with continuous VAEs on datasets 
such as CIFAR-10 \cite{vqvae}.
An extension of this method to conditional supervised generation, out-performs continuous autoencoders on WMT 
English-German translation task \cite{kaiser2018fast}.

\cite{kaiser2018fast} introduced the Latent Transformer, which achieved impressive results using discrete autoencoders for fast neural machine translation. However, additional
training heuristics, namely, exponential moving averages (EMA) of cluster assignment counts, and product quantization \cite{norouzi2013cartesian} were essential to achieve 
competitive results with VQ-VAE. 
In this work, we show that tuning for the code-book size can significantly outperform the results presented in \cite{kaiser2018fast}. We also exploit VQ-VAE's connection with the expectation maximization (EM) 
algorithm 
\cite{dempster1977maximum}, yielding additional improvements. 
With both improvements, 
we achieve a BLEU score of \(22.4\) on English to German translation, outperforming 
\cite{kaiser2018fast} by $2.6$ BLEU. 
Knowledge distillation \cite{hinton2015distilling,seq-d} provides significant gains with our best models and EM, achieving $26.7$ BLEU, which almost matches the autoregressive transformer model with no 
beam search at $27.0$ BLEU, while being \(3.3\times\) faster.

Our contributions can be summarized as follows:
\begin{enumerate}
\item We show that VQ-VAE from \cite{vqvae} can outperform previous state-of-the-art without product quantization.
\item Inspired by the EM algorithm, we introduce a new training algorithm for training discrete variational autoencoders, that outperforms the previous best result with discrete latent autoencoders for neural machine translation.
\item Using EM training, we achieve better image generation results on CIFAR-10,
and with the additional use of knowledge distillation, allows us to develop a non-autoregressive machine translation model
whose accuracy almost matches a strong greedy autoregressive baseline Transformer,
while being \(3.3\) times faster at inference.
\end{enumerate}

\section{VQ-VAE and the Hard EM Algorithm}
\label{sec:connection}

\begin{figure}[!htb]
\centering
\includegraphics[scale=0.37]{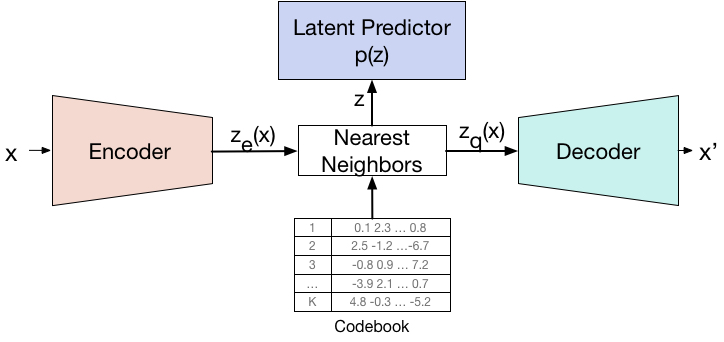}
\caption{VQ-VAE model as described in \cite{vqvae}. 
We use the notation \(x\) to denote the input image, with the 
output of the encoder \(\encoder(\target) \in R^D\) being used to perform nearest neighbor search to select the 
(sequence of) discrete latent variable. The selected discrete latent is used to train the latent predictor model, 
while the embedding \(z_q(x)\) of the selected discrete latent is passed as input to the decoder.}
\end{figure}

The connection between \(K\)-means, and hard EM, or the Viterbi EM algorithm is well known~\cite{bottou1995convergence}, 
where the former can be seen a special
case of hard-EM style algorithm with a
mixture-of-Gaussians model with identity
covariance and uniform prior over cluster
probabilities. In the following sections we briefly explain the VQ-VAE discretization algorithm for completeness and it's connection to classical EM.

\subsection{VQ-VAE discretization algorithm}
\label{sec:vq-vae}
VQ-VAE models the joint distribution  \(P_{\Theta}(x, z)\) where \(\Theta\) are the model parameters, \( x\) is the data point and \( z\) is the sequence of discrete latent variables or codes.
Each position in the encoded sequence has its own set of latent codes.
Given a data point, the discrete latent code in each position is selected independently using the encoder output.
For simplicity, we describe the procedure for selecting the discrete latent code (\(z_i\)) in one position given the data point (\(x_i\)).
The encoder output \(\encoder(\target_i) \in R^D\) is passed through a discretization bottleneck using a nearest-neighbor lookup on embedding vectors \(e \in R^{K \times D}\).
Here \(K\) is the number of latent codes (in a particular position of the discrete latent sequence) in the model.
More specifically, the discrete latent variable assignment is given by, 
\begin{align}\label{eq:nn}
    z_i = \arg\min_{j \in [K]}
    \norm{\encoder(\target_i) - e_j}_2
\end{align}
The selected latent variable's embedding is passed as input to the decoder, 
\begin{align*}
    z_q(\target_i) = e_{z_i}
\end{align*}

The model is trained to minimize: 
\begin{align}\label{eq:beta}
    L = l_r + \beta \norm{\encoder(\target_i) - \sg{z_q(\target_i)}}_2,
\end{align}
where \(l_r\) is the reconstruction loss of the decoder
given \(z_q(\target)\) 
(e.g., the cross entropy loss), and, 
\(\sg{.}\) is the stop gradient operator defined as follows:
\begin{align*}
    \sg{x} = \begin{cases} x \quad \text{forward pass} \\
                           0  \quad \text{backward pass}
             \end{cases}
\end{align*}

It was observed in \cite{kaiser2018fast} that an exponentially moving average (EMA) update of the latent 
embeddings and code-book assignments results in more stable training than using gradient-based methods.

Specifically, they maintain EMA of the following two quantities: 1) the embeddings \(e_j\) for every \(j \in [1, \ldots, K]\) and, 2) the count \(c_j\) measuring the number of encoder hidden states that have \(e_j\) as it's nearest neighbor. The counts are updated in a mini-batch of targets as:
\begin{align}\label{eq:ema1}
    c_j \gets \lambda c_j + (1 - \lambda) \sum_{i} \mathbbm{1}\left[z_q(\target_i) = e_j\right],
\end{align}
with the embedding \(e_j\) being subsequently updated as:
\begin{align}\label{eq:ema2}
    e_j \gets \lambda e_j + (1 - \lambda) \sum_{i} 
    \frac{\mathbbm{1}\left[z_q(\target_i) = e_j\right]z_e(\target_i)}{c_j},
\end{align}
where \(\mathbbm{1}[.]\) is the indicator function and \(\lambda\)
is a decay parameter which we set to \(0.999\) in our experiments. This amounts to doing stochastic gradient in the space of both code-book embeddings and cluster assignments. These techniques have also been successfully 
used in minibatch \(K\)-means \cite{sculley2010web} and online EM \cite{liang2009online,sato2000line}. 

The generative process begins by sampling a sequence of discrete latent codes from an autoregressive model, which we refer to as the Latent Predictor model. 
The decoder then consumes this sequence of discrete latent variables to generate the data.
The autoregressive model which acts as a learned prior is fitted on the discrete latent variables produced by
the encoder.
The architecture of the encoder, the decoder, and the latent predictor model are described in further detail in the 
experiments section.

\subsection{Hard EM and the \(K\)-means algorithm}
\label{sec:hard-em}
In this section we briefly recall the hard Expectation maximization (EM)
algorithm \cite{dempster1977maximum}.
Given a set of data points \((x_1, \dots, x_N)\), the hard EM algorithm approximately solves the following optimization problem:
\begin{align}\label{eq:hard-em}
    \Theta^* = \displaystyle \arg\max_{\Theta} P_{\Theta}(x_1, \dots, x_N) = \displaystyle \arg\max_{\Theta} \max_{z_1, \dots, z_N} P_{\Theta}(x_1, \dots, x_N, z_1, \dots, z_N),
\end{align}
Hard EM performs coordinate descent over the following two coordinates: the model parameters \(\Theta\),
and the hidden variables \(z_1, \dots, z_N\). In other words, hard EM consists of 
repeating the following two steps until convergence:
\begin{enumerate}
    \item \textbf{E step:}  \( (z_1, \dots, z_N) \gets \arg\max_{z_1, \dots, z_N} 
    P_{\Theta}(x_1, \dots, x_N, z_1, \dots, z_N) \),
    \item \textbf{M step:} \( \Theta \gets \arg \max_{\Theta} P_{\Theta}(x_1, \dots, x_N, z_1, \dots, z_N) \)
\end{enumerate}

A special case of the hard EM algorithm is \(K\)-means clustering \cite{macqueen1967some, bottou1995convergence} where the likelihood is modelled by a Gaussian with identity covariance matrix. Here, the means of the \(K\) Gaussians are the parameters to be estimated, 
\begin{align*}
    \Theta = \langle \mu^1, \dots, \mu^K\rangle, \quad \mu^k \in R^D.
\end{align*}
With a uniform prior over the hidden variables (\(P_{\Theta}(z_i) = \frac{1}{K}\)),  
the marginal is given by \(P_{\Theta}(x_i \mid z_i) = 
\mathcal{N}(\mu^{z_i}, I)(x_i)\). In this case, equation~\eqref{eq:hard-em}
is equivalent to:
\begin{align}\label{eq:k-means}
    \left(\mu^1, \dots, \mu^K\right)^* = \arg\max_{\mu^1, \dots, \mu^K} \min_{z_1, \dots, z_N}
    \sum_{i=1}^N \norm{\mu^{z_i} - x_i}^2_2
\end{align}

Note that optimizing equation~\eqref{eq:k-means} is NP-hard, however one can find a
local optima by applying coordinate descent until convergence:
\begin{enumerate}
\item \textbf{E step:} Cluster assignment is given by, \begin{align}\label{eq:hard-em-e} z_i \gets \arg\min_{j \in [K]} \norm{\mu^{j} - x_i}^2_2, \end{align}
\item \textbf{M step:} The means of the clusters are updated as, \begin{align}\label{eq:hard-em-m} c_j \gets \sum_{i=1}^N \mathbbm{1}[z_i = j] ; \quad
\mu^j \gets \frac{1}{c_j}\sum_{i=1}^N \mathbbm{1}[z_i = j] x_i. \end{align}
\end{enumerate}
We can now easily see the connections between the training updates of VQ-VAE and \(K\)-means clustering.
The encoder output \(\encoder(\target) \in R^D\) corresponds to the data point while the discrete latent variables corresponds to clusters.
Given this, Equation~\ref{eq:nn} is equivalent to the E-step (Equation~\ref{eq:hard-em-e}) and the EMA updates in 
Equation~\ref{eq:ema1} and Equation~\ref{eq:ema2} converge to the M-step (Equation ~\ref{eq:hard-em-m}) in the 
limit.
The M-step in \(K\)-means overwrites the old values while the EMA updates interpolate between the old 
values and the M step update.

\section{VQ-VAE training with EM}
\label{sec:em}
In this section, we investigate a new training strategy for VQ-VAE using the soft EM algorithm.
\subsection{Soft EM}
First, we briefly describe the soft EM algorithm. While the hard EM procedure selects one cluster or latent variable assignment for a data point, here the data point is assigned to a mixture of clusters. Now, the optimization objective is given by, 
\begin{align*}
    \Theta^* &= \arg\max_{\Theta} P_{\Theta} (x_1, \dots, x_N) \\
             &= \arg\max_{\Theta} \sum_{z_1, \dots, z_N} P_\Theta(x_1, \dots, x_N, z_1, \dots, z_N)
\end{align*}
Coordinate descent algorithm is again used to approximately solve the above optimization algorithm. The E and M step are given by:
\begin{enumerate}
    \item \textbf{E step:}  \begin{align}\label{eq:soft-em-e} \rho(z_i) \gets P_\Theta(z_i \mid x_i), \end{align} 
    \item \textbf{M step:} \begin{align}\label{eq:soft-em-m} \Theta \gets \arg\max_{\Theta} \mathbbm{E}_{z_i \sim \rho}[\log{P_\Theta(x_i, z_i)}] \end{align}
\end{enumerate}

\subsection{Vector Quantized Autoencoders trained with EM}
\label{sec:soft-em}
Now, we describe vector quantized autoencoders training using the soft EM algorithm.
As discussed in the previous section, the encoder output \(\encoder(\target) \in R^D\) corresponds to the data point while the discrete latent variables corresponds to clusters.
The E step instead of hard assignment now produces a probability distribution over the set of discrete latent 
variables (Equation~\ref{eq:soft-em-e}). Following VQ-VAE, we continue to assume a uniform prior over clusters, 
since we observe that training the cluster priors seemed to cause the cluster assignments to collapse to only a 
few clusters. 
The probability distribution is modeled as a Gaussian with identity covariance matrix, 
         \begin{align*} P_\Theta(z_i \mid \encoder(\target_i))
\propto e^{-\norm{e_{z_i} - \encoder(\target_i)}^2_2}
         \end{align*}
Since computing the expectation in the M step (Equation~\ref{eq:soft-em-m}) is computationally infeasible in our case, we instead perform Monte-Carlo Expectation Maximization \cite{wei1990monte} by drawing \(m\) samples 
\(z_i^1, \cdots, z_i^m \sim \multinomial \left(- \norm{e_1 - \encoder(\target_i)}^2_2, \dots, - \norm{e_K - \encoder(\target_i)}^2_2\right)\), where \(\multinomial(l_1, \dots, l_K)\) refers to the \(K\)-way multinomial distribution with 
logits \(l_1, \dots, l_K\). 
Thus, the E step can be finally written as:  \\
\begin{align*}
\textbf{E step:} \qquad 
        z_i^1, \dots, z_i^m \gets \multinomial \left(- \norm{e_1 - \encoder(\target_i)}^2_2, \dots, - \norm{e_K - \encoder(\target_i)}^2_2\right)
\end{align*}

The model parameters \(\Theta\) are then updated to maximize this Monte-Carlo estimate in the M step given by \\
\begin{align*}
 \textbf{M step:} \qquad c_j \gets \frac{1}{m} \sum_{i=1}^N \sum_{l=1}^m \mathbbm{1}\left[z^l_i = j\right]; \qquad
e_j \gets \frac{1}{m c_j}\sum_{i=1}^N \sum_{l=1}^m \mathbbm{1}\left[z^l_i = j\right] \encoder(x_i).
\end{align*}
Instead of exactly following the above M step update, we use the EMA version of this update similar to the one described in Section~\ref{sec:vq-vae}. 

When sending the embedding of the discrete latent to the decoder, instead of sending the posterior mode,
\(\argmax_{z} P(z \mid x)\), similar to hard EM and \(K\)-means, we send the average of the embeddings of the sampled latents:
\begin{align}
    z_q(x_i) = \frac{1}{m} \sum_{l=1}^m e_{z^l_i}.
\end{align}
Since \(m\) latent code embeddings are sent to the decoder in the forward pass, all of them are updated in the backward pass for a single training example. In hard EM training, only one of them is updated during training. Sending averaged embeddings also results in more stable training using the soft EM algorithm compared to hard EM as shown in Section~\ref{sec:experiments}.

To train the latent predictor model (Section~\ref{sec:vq-vae}) in this case, we use an approach similar to \emph{label smoothing}
\cite{pereyra2017regularizing}: the latent predictor model is trained to minimize
the cross entropy loss with the labels being the average of the one-hot labels 
of \(z^1_i, \dots, z^m_i\).

\section{Other Related Work}

Variational autoencoders were first introduced by \cite{kingma2016improved, rezende2014stochastic}
for training continuous representations; unfortunately, training them for discrete latent variable models 
has proved challenging. 
One promising approach has been to use various gradient estimators for discrete latent variable models, starting with the REINFORCE estimator of \cite{williams1992simple}, an unbiased, high-variance gradient estimator. 
An alternate approach towards gradient estimators is to use continuous
relaxations of categorical distributions, for e.g., the Gumbel-Softmax reparametrization trick
\cite{gs1, gs2}. These methods provide biased but low variance gradients for training.

Machine translation using deep neural networks have been shown to achieve impressive results  
\cite{sutskever14,bahdanau2014neural,cho2014learning,transformer}.
The state-of-the-art models in Neural Machine Translation are all auto-regressive, which means that during decoding, the model consumes all previously generated tokens to predict the next one. 
Very recently, there have been multiple efforts to speed-up machine translation decoding.
\cite{nonautoregnmt} attempts to address this issue by using the Transformer model~\cite{transformer} together with the REINFORCE algorithm \cite{williams1992simple}, to model the \emph{fertilities} of words. 
The main drawback of the approach of \cite{nonautoregnmt} is the need for extensive fine-tuning to make policy gradients work, as well as the non-generic nature of the solution. 
\cite{lee2018deterministic} propose a non-autoregressive model using iterative refinement. 
Here, instead of decoding the target sentence in one-shot, the output is successively refined to produce the final output. 
While the output is produced in parallel at each step, the refinement steps happen sequentially.


\section{Experiments}
\label{sec:experiments}
We evaluate our proposed methods on unconditional image generation on the CIFAR-10 dataset and supervised conditional language generation on the WMT English-to-German translation task. 
Our models and generative process follow the architecture proposed in \cite{vqvae} for unconditional image generation, and \cite{kaiser2018fast} for neural machine translation.
For all our experiments, we use the Adam~\cite{kingma2014adam} optimizer and decay the learning rate exponentially after initial warm-up steps. Unless
otherwise stated, the dimension of the hidden states of the encoder and the decoder is \(512\), see Table~\ref{tab:model-size} for a comparison
of models with lower dimension.
The code to reproduce our 
experiments will be released with the next version of the paper.

\subsection{Machine Translation}
\begin{figure}[!htb]
\centering
\includegraphics[scale=0.37]{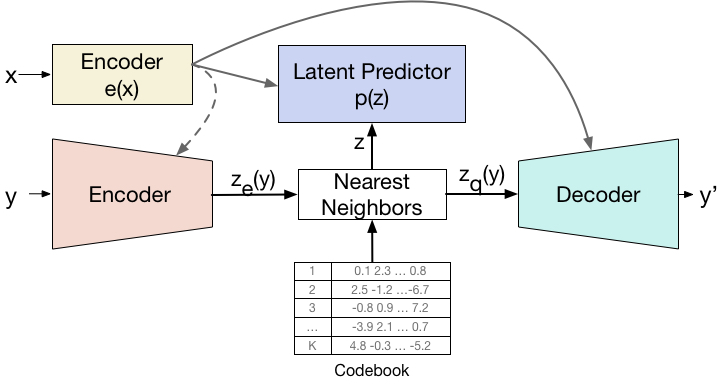}
\caption{VQ-VAE model adapted to conditional supervised translation as described in \cite{kaiser2018fast}.
We use \(x\) and \(y\) to denote the source and target sentence respectively. 
The encoder, the decoder and the latent predictor now additionally condition on the source sentence \(x\).}
\label{fig:translation}
\end{figure}

In Neural Machine Translation with latent variables, we model $P(y, z \mid x)$, where $y$ and $x$ are the target and source sentence respectively. 
Our model architecture, depicted in Figure~\ref{fig:translation}, is similar to the one in \cite{kaiser2018fast}.
The encoder function is a series of strided convolutional layers with residual convolutional layers in between and takes target sentence $y$ as input.
The source sentence \(x\) is converted to a sequence of hidden states through multiple causal self-attention layers.
In \cite{kaiser2018fast}, the encoder of the autoencoder  attends additionally to this sequence of continuous representation of the source sentence. We use VQ-VAE as the 
discretization algorithm. The decoders, applied after the bottleneck layer uses transposed convolution layers whose continuous output is fed to a transformer decoder with 
causal attention, which generates the output. 

The results are summarized in Table~\ref{tab:nmt}. Our implementation of VQ-VAE achieves a significantly better BLEU score and faster decoding speed compared to \cite{kaiser2018fast}.
We found that tuning the code-book size (number of clusters) for using $2^{12}$ discrete latents achieves the best accuracy which is 16 times smaller as compared to the code-book size in \cite{kaiser2018fast}.
Additionally, we see a large improvement in the performance of the model by using sequence-level distillation \cite{seq-d}, as has been observed previously in non-autoregressive models \cite{nonautoregnmt,lee2018deterministic}.
Our teacher model is a base Transformer \cite{transformer} that achieves a BLEU score of $28.1$ and $27.0$ on the WMT'14 test set using beam search decoding and greedy decoding respectively. For distillation purposes, we use the beam search decoded Transformer.
Our VQ-VAE model trained with soft EM and distillation, achieves a BLEU score of $26.7$, without noisy parallel decoding~\cite{nonautoregnmt}. 
This perforamce is $1.4$ bleu points lower than an autoregressive model decoded with a beam size of $4$, while being $4.1 \times$ faster. 
Importantly, we nearly match the same autoregressive model with beam size $1$, with a $3.3 \times$ speedup.

The length of the sequence of discrete latent variables is shorter than that of target sentence $y$. 
Specifically, at each compression step of the encoder we reduce its length by half. We denote by 
\(n_c\), the compression 
factor for the latents, i.e. the number of steps for which we do this compression. In almost all our 
experiments, we use \(n_c=3\) reducing the length by 8. We can decrease the decoding time further by increasing 
the number of compression steps. As shown in Table \ref{tab:nmt}, by setting \(n_c\) to 4, the decoding time 
drops to 58 milliseconds achieving 25.4 BLEU while a NAT model with similar decoding speed achieves only 18.7 
BLEU. Note that, all NAT models also train with sequence level knowledge distillation from an autoregressive 
teacher.  

\subsubsection{Analysis}

\begin{figure}[!htb]
    \centering
    \includegraphics[scale=.32]{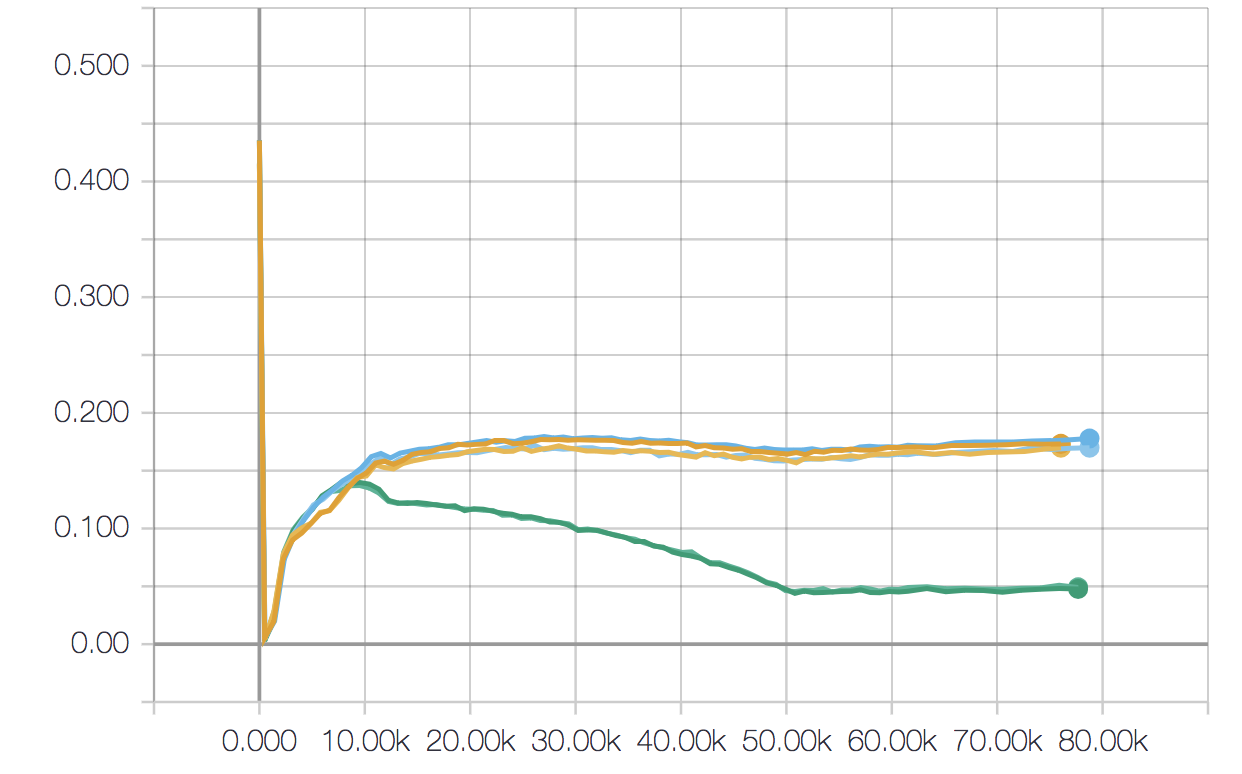}
    \caption{Comparison of hard EM (green curve) vs soft EM   with different number of samples (yellow and blue curves) on the WMT'14 English-German translation dataset with a code-book
    size of \(2^{14}\), with the encoder of the discrete autoencoder attending to the output of the 
    encoder of the source sentence as in \cite{kaiser2018fast}. 
    The \(y\)-axis denotes the teacher-forced BLEU score on the test set. 
    Notice that the hard EM/\(K\)-means run collapsed, while the soft EM runs exhibit more stability.} 
    \label{fig:emvshard}
\end{figure}

\paragraph{Attention to Source Sentence Encoder:} While the encoder of the discrete autoencoder  in \cite{kaiser2018fast} attends to the output of the encoder of the source sentence, we find that to be unnecessary, with both models achieving the same BLEU score with \(2^{12}\) latents. Also, removing this attention step results in more stable training particularly for large code-book sizes, see e.g., Figure~\ref{fig:emvshard}.   

\paragraph{VQ-VAE vs Other Discretization Techniques: } 
We compare the Gumbel-Softmax of \cite{gs1, gs2} and the improved semantic hashing discretization technique proposed in \cite{kaiser2018fast} to VQ-VAE.
When trained with sequence level knowledge distillation, the model using 
Gumbel-Softmax 
reached \(23.2\) BLEU, the model using improved semantic hashing reached 
\(24.1\) BLEU,
while the model using VQ-VAE reached \(26.4\) BLEU on WMT'14 English-German.

\paragraph{Size of Discrete Latent Variable code-book:} Table \ref{tab:z-size} in Appendix shows the BLEU score for different code-book sizes for models trained using hard EM without distillation. While \cite{kaiser2018fast} use \(2^{16}\) as their code-book size, we find that \(2^{12}\) gives the best performance.

\paragraph{Robustness of EM to Hyperparameters:} While the soft EM training gives a small performance improvement, we find that it also leads to more robust training (Figure \ref{fig:emvshard}).

\paragraph{Model Size:} The effect of model size on BLEU score for models trained with soft EM and distillation is shown in Table \ref{tab:model-size} in Appendix.

\paragraph{Number of samples in Monte-Carlo EM update}
While training with soft EM, we perform a Monte-Carlo update with a small number of samples (Section \ref{sec:soft-em}). Table \ref{tab:em} in Appendix shows the impact of number of samples on the final BLEU score.
\begin{table}
\centering
\begin{threeparttable}
\begin{tabular}{l?c|c|c|c|c}
\toprule
{Model} & \(n_c\) & \(n_s\) & BLEU & {Latency} & Speedup \\ 
\midrule
Autoregressive Model (beam size=4) & - & - & 28.1 & \(331\) ms & \(1 \times\)   \\
Autoregressive Baseline (no beam-search) & - & - & 27.0 & 265 ms & \(1.25\times\)  \\
\midrule
NAT + distillation                & - & - & 17.7   & 39 ms & \(15.6\times\) \tnote{*}\\
NAT + distillation + NPD=10            & -    & - & 18.7  & 79 ms & \(7.68\times\) \tnote{*} \\
NAT + distillation + NPD=100            & -   & - & 19.2  &  257 ms & \(2.36\times\)  \tnote{*}\\
LT + Semhash & - & - & 19.8 & 105 ms & \(3.15\times\)\\
\hline
\multicolumn{5}{c}{Our Results} \\
\midrule
VQ-VAE & 3 & - & 21.4  & 81 ms & \(4.08\times\) \\
VQ-VAE with EM & 3 & 5 &  22.4 & 81 ms & \(4.08\times\)\\
\hline
VQ-VAE + distillation & 3 & - & 26.4  & 81 ms & \(4.08\times\) \\
VQ-VAE with EM + distillation & 3 & 10 & \textbf{26.7} & 81 ms & \(4.08\times\) \\
VQ-VAE with EM + distillation & 4 & 10 &  25.4 & 58 ms & \(5.71\times\)\\
\bottomrule
\end{tabular}
\vspace{1mm}
\caption{BLEU score and decoding times for different models on the WMT'14 English-German translation dataset.
The baseline is the autoregressive Transformer of \cite{transformer} with no beam search,
NAT denotes the Non-Autoregressive Transformer of \cite{nonautoregnmt}, and LT + Semhash denotes the Latent Transformer 
from \cite{vqvae} using the improved semantic hashing discretization technique of\cite{isemhash}. NPD is noisy parallel decoding as described in \cite{nonautoregnmt}. 
We use the notation \(n_c\) to denote
the compression factor for the latents, and the notation \(n_s\) to denote the number of samples used to perform
the Monte-Carlo approximation of the EM algorithm. Distillation refers to sequence level knowledge distillation 
from \cite{seq-d}.
We used a code-book of size \(2^{12}\) for EM and
decoding is performed on a single CPU machine with an NVIDIA GeForce
GTX 1080 with a batch size of \(1\)}
\label{tab:nmt}
\begin{tablenotes}
\item[*] \footnotesize{Speedup reported for these items are compared to the
decode time of \(408\) ms for an autoregressive Transformer from \cite{nonautoregnmt}. }
\end{tablenotes}
\end{threeparttable}
\end{table}

\subsection{Image Generation}

\begin{figure}[!httb]
    \includegraphics[scale=0.75]{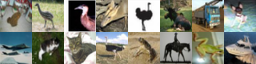}
    \hspace{2mm}
    \includegraphics[scale=0.75]{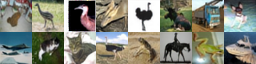}
    \caption{Samples of original and reconstructed images from CIFAR-10 using VQ-VAE trained using
    EM with a code-book of size \(2^{8}\).} 
    \label{fig:reconst}
\end{figure}

\begin{table}[!htb]\label{table:cifar}
\begin{center}
\renewcommand*{\arraystretch}{1}
\begin{tabular}{l?c|c}
\toprule
{Model}   & \(n_s\) & Log perplexity \\
\midrule
ImageTransformer & - & $2.92$ \\
VAE & - & $4.51$ \\
VQ-VAE \cite{vqvae} & - & $\mathbf{4.67}$ \\
\midrule
VQ-VAE (Ours)  & - & $4.83$ \\
EM & $5$ & $\mathbf{4.80}$ \\
\bottomrule
\end{tabular}
\end{center}
\caption{Log perplexity on CIFAR-10 measured in bits/dim. 
We train our
VQ-VAE models on a field of \(8 \times 8 \times 10\) latents
with a code-book of size \(2^{8}\), while VQ-VAE refers to the
results from \cite{vqvae} which was trained on a field of 
\(8 \times 8 \times 10\) latents on a code-book of size \(2^9\). Note that, VQ-VAE \cite{vqvae}
takes a unigram prior for each latent in the sequence independently instead of log-perplexity from the Latent Predictor model.} 
\end{table}

We train the unconditional VQ-VAE model on the CIFAR-10 data set, modeling the joint probability \(P(x, z)\), where
\(x\) is the image and \(z\) are the discrete latent codes. We use a field of
\(8 \times 8\ \times 10\) latents with a code-book of size \(2^{8}\) each containing $512$ dimensions.
We maintain the same encoder and decoder as used in Machine Translation. Our Latent Predictor, uses an Image 
Transformer \cite{imagetrans} auto-regressive decoder with $6$ layers of local $1D$ self-attention. For the 
encoder, we use $4$ convolutional layers, with kernel size $5\times 5$ and strides 
$2\times 2$, followed by $2$ residual 
layers, and a single dense layer. For the decoder, we use a single dense layers, $2$ residual layers, and $4$ 
deconvolutional layers.

We calculate the lower bound on negative log-likelihood in terms of the Latent Predictor loss \(l_{lp}\)
and the negative log-perplexity \(l_p\) of the autoencoder.
Let \(n_x\) be the total number of positions in the image, and \(n_z\) the number of latent codes. 
Then the lower-bound on the 
negative log-likelihood 
\(
    - \log{P(x)} = - \log{P(x \mid z)} - \log{P(z)},
\)
is computed in bits/dim as
\(
   \left(\frac{l_p * {n_x} + l_{lp} * {n_z}}{n_x}\right) * \log_2{e}.
   \)
Note that for CIFAR-10, \(n_x = 32 \times 32 \times 3\) while \(n_z = 8 \times 8 \times 10\).
We report the results in Table~\ref{table:cifar} and show reconstructions from the autoencoder in 
Figure~\ref{fig:reconst}. As seen from the results, our VQ-VAE model with EM gets \(0.03\) bits/dim better negative log-likelihood as compared to the baseline VQ-VAE.

\section{Conclusion}
We investigate an alternate training technique for VQ-VAE inspired by its connection to the EM algorithm. 
Training the discrete bottleneck with EM helps us achieve better image generation results on CIFAR-10, and 
together with knowledge distillation, allows us to develop a non-autoregressive machine translation model
whose accuracy almost matches the greedy autoregressive baseline,
while being 3.3 times faster at inference.
\bibliography{deeplearn}

\begin{thebibliography}{10}

\bibitem{bahdanau2014neural}
Dzmitry Bahdanau, Kyunghyun Cho, and Yoshua Bengio.
\newblock Neural machine translation by jointly learning to align and
  translate.
\newblock {\em CoRR}, abs/1409.0473, 2014.

\bibitem{bottou1995convergence}
Leon Bottou and Yoshua Bengio.
\newblock Convergence properties of the k-means algorithms.
\newblock In {\em Advances in neural information processing systems}, pages
  585--592, 1995.

\bibitem{cho2014learning}
Kyunghyun Cho, Bart van Merrienboer, Caglar Gulcehre, Fethi Bougares, Holger
  Schwenk, and Yoshua Bengio.
\newblock Learning phrase representations using {RNN} encoder-decoder for
  statistical machine translation.
\newblock {\em CoRR}, abs/1406.1078, 2014.

\bibitem{dempster1977maximum}
Arthur~P Dempster, Nan~M Laird, and Donald~B Rubin.
\newblock Maximum likelihood from incomplete data via the em algorithm.
\newblock {\em Journal of the royal statistical society. Series B
  (methodological)}, pages 1--38, 1977.

\bibitem{goodfellow2014generative}
Ian Goodfellow, Jean Pouget-Abadie, Mehdi Mirza, Bing Xu, David Warde-Farley,
  Sherjil Ozair, Aaron Courville, and Yoshua Bengio.
\newblock Generative adversarial nets.
\newblock In {\em Advances in neural information processing systems}, pages
  2672--2680, 2014.

\bibitem{nonautoregnmt}
Jiatao Gu, James Bradbury, Caiming Xiong, Victor~O.K. Li, and Richard Socher.
\newblock Non-autoregressive neural machine translation.
\newblock {\em CoRR}, abs/1711.02281, 2017.

\bibitem{hinton2015distilling}
Geoffrey Hinton, Oriol Vinyals, and Jeff Dean.
\newblock Distilling the knowledge in a neural network.
\newblock {\em arXiv preprint arXiv:1503.02531}, 2015.

\bibitem{gs1}
Eric Jang, Shixiang Gu, and Ben Poole.
\newblock Categorical reparameterization with gumbel-softmax.
\newblock {\em CoRR}, abs/1611.01144, 2016.

\bibitem{isemhash}
\L{}ukasz Kaiser and Samy Bengio.
\newblock Discrete autoencoders for sequence models.
\newblock {\em CoRR}, abs/1801.09797, 2018.

\bibitem{kaiser2018fast}
{\L}ukasz Kaiser, Aurko Roy, Ashish Vaswani, Niki Pamar, Samy Bengio, Jakob
  Uszkoreit, and Noam Shazeer.
\newblock Fast decoding in sequence models using discrete latent variables.
\newblock {\em arXiv preprint arXiv:1803.03382}, 2018.

\bibitem{kalchbrenner2016video}
Nal Kalchbrenner, Aaron van~den Oord, Karen Simonyan, Ivo Danihelka, Oriol
  Vinyals, Alex Graves, and Koray Kavukcuoglu.
\newblock Video pixel networks.
\newblock {\em arXiv preprint arXiv:1610.00527}, 2016.

\bibitem{seq-d}
Yoon Kim and Alexander Rush.
\newblock Sequence-level knowledge distillation.
\newblock 2016.

\bibitem{kingma2014adam}
Diederik~P Kingma and Jimmy Ba.
\newblock Adam: A method for stochastic optimization.
\newblock {\em arXiv preprint arXiv:1412.6980}, 2014.

\bibitem{kingma2016improved}
Diederik~P Kingma, Tim Salimans, Rafal Jozefowicz, Xi~Chen, Ilya Sutskever, and
  Max Welling.
\newblock Improved variational inference with inverse autoregressive flow.
\newblock In {\em Advances in Neural Information Processing Systems}, pages
  4743--4751, 2016.

\bibitem{lee2018deterministic}
Jason Lee, Elman Mansimov, and Kyunghyun Cho.
\newblock Deterministic non-autoregressive neural sequence modeling by
  iterative refinement.
\newblock {\em arXiv preprint arXiv:1802.06901}, 2018.

\bibitem{liang2009online}
Percy Liang and Dan Klein.
\newblock Online em for unsupervised models.
\newblock In {\em Proceedings of human language technologies: The 2009 annual
  conference of the North American chapter of the association for computational
  linguistics}, pages 611--619. Association for Computational Linguistics,
  2009.

\bibitem{macqueen1967some}
James MacQueen et~al.
\newblock Some methods for classification and analysis of multivariate
  observations.
\newblock In {\em Proceedings of the fifth Berkeley symposium on mathematical
  statistics and probability}, volume~1, pages 281--297. Oakland, CA, USA,
  1967.

\bibitem{gs2}
Chris~J. Maddison, Andriy Mnih, and Yee~Whye Teh.
\newblock The concrete distribution: {A} continuous relaxation of discrete
  random variables.
\newblock {\em CoRR}, abs/1611.00712, 2016.

\bibitem{norouzi2013cartesian}
Mohammad Norouzi and David~J Fleet.
\newblock Cartesian k-means.
\newblock In {\em Computer Vision and Pattern Recognition (CVPR), 2013 IEEE
  Conference on}, pages 3017--3024. IEEE, 2013.

\bibitem{imagetrans}
Niki Parmar, Ashish Vaswani, Jakob Uszkoreit, Lukasz Kaiser, Noam Shazeer, and
  Alexander Ku.
\newblock Image transformer.
\newblock {\em arXiv}, 2018.

\bibitem{pereyra2017regularizing}
Gabriel Pereyra, George Tucker, Jan Chorowski, {\L}ukasz Kaiser, and Geoffrey
  Hinton.
\newblock Regularizing neural networks by penalizing confident output
  distributions.
\newblock {\em arXiv preprint arXiv:1701.06548}, 2017.

\bibitem{reed2017parallel}
Scott Reed, A{\"a}ron van~den Oord, Nal Kalchbrenner, Sergio~G{\'o}mez
  Colmenarejo, Ziyu Wang, Dan Belov, and Nando de~Freitas.
\newblock Parallel multiscale autoregressive density estimation.
\newblock {\em arXiv preprint arXiv:1703.03664}, 2017.

\bibitem{rezende2014stochastic}
Danilo~Jimenez Rezende, Shakir Mohamed, and Daan Wierstra.
\newblock Stochastic backpropagation and approximate inference in deep
  generative models.
\newblock {\em CoRR}, abs/1401.4082, 2014.

\bibitem{salimans2017pixelcnn++}
Tim Salimans, Andrej Karpathy, Xi~Chen, and Diederik~P Kingma.
\newblock Pixelcnn++: Improving the pixelcnn with discretized logistic mixture
  likelihood and other modifications.
\newblock {\em arXiv preprint arXiv:1701.05517}, 2017.

\bibitem{sato2000line}
Masa-Aki Sato and Shin Ishii.
\newblock On-line em algorithm for the normalized gaussian network.
\newblock {\em Neural computation}, 12(2):407--432, 2000.

\bibitem{sculley2010web}
David Sculley.
\newblock Web-scale k-means clustering.
\newblock In {\em Proceedings of the 19th international conference on World
  wide web}, pages 1177--1178. ACM, 2010.

\bibitem{sutskever14}
Ilya Sutskever, Oriol Vinyals, and Quoc~V. Le.
\newblock Sequence to sequence learning with neural networks.
\newblock In {\em Advances in Neural Information Processing Systems}, pages
  3104--3112, 2014.

\bibitem{theis2017lossy}
Lucas Theis, Wenzhe Shi, Andrew Cunningham, and Ferenc Husz{\'a}r.
\newblock Lossy image compression with compressive autoencoders.
\newblock {\em arXiv preprint arXiv:1703.00395}, 2017.

\bibitem{van2016wavenet}
Aaron Van Den~Oord, Sander Dieleman, Heiga Zen, Karen Simonyan, Oriol Vinyals,
  Alex Graves, Nal Kalchbrenner, Andrew Senior, and Koray Kavukcuoglu.
\newblock Wavenet: A generative model for raw audio.
\newblock {\em arXiv preprint arXiv:1609.03499}, 2016.

\bibitem{van2016conditional}
Aaron van~den Oord, Nal Kalchbrenner, Lasse Espeholt, Oriol Vinyals, Alex
  Graves, et~al.
\newblock Conditional image generation with pixelcnn decoders.
\newblock In {\em Advances in Neural Information Processing Systems}, pages
  4790--4798, 2016.

\bibitem{vqvae}
A{\"{a}}ron van~den Oord, Oriol Vinyals, and Koray Kavukcuoglu.
\newblock Neural discrete representation learning.
\newblock {\em CoRR}, abs/1711.00937, 2017.

\bibitem{transformer}
Ashish Vaswani, Noam Shazeer, Niki Parmar, Jakob Uszkoreit, Llion Jones,
  Aidan~N. Gomez, Lukasz Kaiser, and Illia Polosukhin.
\newblock Attention is all you need.
\newblock {\em CoRR}, 2017.

\bibitem{vinyals2015show}
Oriol Vinyals, Alexander Toshev, Samy Bengio, and Dumitru Erhan.
\newblock Show and tell: A neural image caption generator.
\newblock In {\em Computer Vision and Pattern Recognition (CVPR), 2015 IEEE
  Conference on}, pages 3156--3164. IEEE, 2015.

\bibitem{wei1990monte}
Greg~CG Wei and Martin~A Tanner.
\newblock A monte carlo implementation of the em algorithm and the poor man's
  data augmentation algorithms.
\newblock {\em Journal of the American statistical Association},
  85(411):699--704, 1990.

\bibitem{williams1992simple}
Ronald~J Williams.
\newblock Simple statistical gradient-following algorithms for connectionist
  reinforcement learning.
\newblock In {\em Reinforcement Learning}, pages 5--32. Springer, 1992.

\end{thebibliography}
\bibliographystyle{plain}
\clearpage

\appendix
\section{Ablation Tables}
\renewcommand*{\arraystretch}{1.}
\begin{table}[H]
\begin{center}
\begin{tabular}{l?c|c}
\toprule
{\bf Model} & Code-book size & BLEU  \\ 
\midrule
VQ-VAE  & \(2^{10}\) & 20.8 \\
VQ-VAE & \(2^{12}\) & 21.6 \\
VQ-VAE & \(2^{14}\)  & 21.0 \\
VQ-VAE & \(2^{16}\)  & 21.8 \\
\bottomrule
\end{tabular}
\vspace{1mm}
\caption{Results showing the impact of code-book size on BLEU score.}
\label{tab:z-size}
\end{center}
\end{table}

\renewcommand*{\arraystretch}{1.}
\begin{table}[H]
\begin{center}
\begin{tabular}{l?c|c|c|c|c}
\toprule
{\bf Model} & \(n_c\)  & \(n_s\) & BLEU & Latency & Speedup \\ 
\midrule
VQ-VAE + distillation & 3 & - & 26.4  & 81 ms & 4.08\(\times\) \\
VQ-VAE with EM + distillation & 3 & 5 &  26.4 & 81 ms & 4.08\(\times\)  \\
VQ-VAE with EM + distillation & 3 & 10 & \textbf{26.7} & 81 ms & 4.08\(\times\)  \\
VQ-VAE with EM + distillation & 3 & 25 & 26.6 & 81 ms & 4.08\(\times\) \\
VQ-VAE with EM + distillation & 3 & 50 & 26.5 & 81 ms & 4.08\(\times\) \\
\hline
VQ-VAE + distillation & 4 & - & 22.4 & 58 ms & 5.71\(\times\)  \\
VQ-VAE with EM + distillation & 4 & 5 & 22.3 & 58 ms & 5.71\(\times\)  \\
VQ-VAE with EM + distillation & 4 & 10 & 25.4 & 58 ms & 5.71\(\times\)  \\
VQ-VAE with EM + distillation & 4 & 25 & 25.1 & 58 ms & 5.71\(\times\) \\
VQ-VAE with EM + distillation & 4 & 50 & 23.6 & 58 ms & 5.71\(\times\) \\
\bottomrule
\end{tabular}
\vspace{1mm}
\caption{Results showing the impact of number of samples used to perform the Monte-Carlo EM update on the BLEU score.}
\label{tab:em}
\end{center}
\end{table}

\begin{table}[H]
\begin{center}
\begin{tabular}{l?c|c|c|c|c}
\toprule
{ Model} & Hidden Vector dimension & \(n_s\) & BLEU & Latency & Speedup \\ 
\midrule
VQ-VAE + distillation & 256 & - & 24.5 & 76 ms & \(4.36\times\) \\
VQ-VAE with EM + distillation & 256 & 10 & 21.9 & 76 ms & \(4.36\times\) \\
VQ-VAE with EM + distillation & 256 & 25 & 25.8 & 76 ms & \(4.36\times\) \\
\hline
VQ-VAE + distillation &  384 & - & 25.6 &  80 ms & \(4.14\times\) \\
VQ-VAE with EM + distillation &  384 & 10 & 22.2 & 80 ms & \(4.14\times\) \\
VQ-VAE with EM + distillation &  384 & 25 & 26.2  & 80 ms & \(4.14\times\)\\
\bottomrule
\end{tabular}
\vspace{1mm}
\caption{Results showing the impact of the dimension of 
the word embeddings and the hidden layers of the model.}
\label{tab:model-size}
\end{center}
\end{table}

\end{document}